\documentclass{article}
\usepackage{spconf,amsmath,graphicx}
\usepackage{stfloats}
\usepackage{booktabs}
\usepackage{float}
\usepackage{graphicx}
\usepackage{cite}
\usepackage{picinpar}
\usepackage{amsmath}
\usepackage{url}
\usepackage{flushend}
\usepackage[utf8]{inputenc}
\usepackage{subcaption}
\usepackage{colortbl}
\usepackage{soul}
\usepackage{multirow}
\usepackage{fancyhdr}
\usepackage{pifont}
\usepackage{color}
\usepackage{alltt}
\usepackage[hidelinks]{hyperref}
\usepackage{enumerate}
\usepackage{siunitx}
\usepackage{breakurl}
\usepackage{epstopdf}
\usepackage{pbox}
\usepackage[marginal]{footmisc}
\renewcommand{\thefootnote}{}
\usepackage{bbding}
\begin{document}

\title{Power-LLaVA: Large Language and Vision Assistant \\ for Power Transmission line Inspection}

\name{\parbox{\linewidth}{\centering Jiahao Wang$^{*}$ \qquad Mingxuan Li$^{*}$ \qquad Haichen Luo\qquad Jinguo Zhu\\ Aijun Yang\qquad Mingzhe Rong \qquad Xiaohua Wang\textsuperscript{\Envelope}  }}
\address{State Key Laboratory of Electrical Insulation and Power Equipment \\ Xi’an Jiaotong University, China}
\maketitle

\thispagestyle{fancy}
\lfoot{© 2024 IEEE. Personal use of this material is permitted. Permission from IEEE must be obtained for all other uses, in any current or future media, including reprinting/republishing this material for advertising or promotional purposes, creating new collective works, for resale or redistribution to servers or lists, or reuse of any copyrighted component of this work in other works.}
\cfoot{}
\rhead{}
\renewcommand{\headrulewidth}{0mm}

\begin{abstract}
The inspection of power transmission line  has achieved notable achievements  in the past few years, primarily due to the integration of deep learning technology. However, current inspection approaches continue to encounter difficulties in generalization and intelligence, which restricts their further applicability.
In this paper, we introduce Power-LLaVA, the first large language and vision  assistant designed to offer  professional and reliable inspection services  for power transmission line by engaging in dialogues with humans.
Moreover, we also construct a large-scale and high-quality dataset specialized for the inspection task.
By employing a  two-stage training strategy on the  constructed dataset, Power-LLaVA demonstrates exceptional  performance at a comparatively low training cost.
Extensive experiments further prove the great capabilities of Power-LLaVA within the realm of power transmission line inspection.
Code shall be released.

\end{abstract}
\begin{keywords}
Power transmission line inspection, Large language-vision assistant, Two-stage training strategy
\end{keywords}

\section{Introduction}
\renewcommand{\thefootnote}{\fnsymbol{footnote}}
\footnotetext[1]{Equal contribution. 
 \textsuperscript{\Envelope} Corresponding author: Xiaohua Wang $<$\url{xhw@mail.xjtu.edu.cn}$>$
}
In the past few years, significant advancements have been seen in power transmission line inspection, largely due to the incorporation of deep learning techniques \cite{electronics9061030,li2023df,10129830}. For instance, DF-YOLO \cite{li2023df} incorporates Deformable Convolutions (DCN) and the SimAM attention mechanism to enhance the performance of original YOLOv7-tiny \cite{wang2023yolov7}, effectively boosting the accuracy of foreign object detection for transmission line. GA-Net \cite{10129830} uses Genetic Algorithm (GA) and  Space-to-Depth (SPD) convolution methods to improve the original YOLOv7 \cite{wang2023yolov7}, increasing accuracy and accelerating the convergence speed effectively. Nevertheless, current methods exhibit limitations in their generalization and intelligence, resulting in suboptimal performance when confronted with diverse and complex application scenarios \cite{momoh2018electric}. 

The remarkable success of Large Language Models (LLMs) \cite{ouyang2022training,touvron2023llama,taori2023alpaca} has recently inspired the development of Large Multimodal  Models (LMMs) that extend  beyond LLMs, as examplified by LLaVA \cite{liu2023visual}, Qwen \cite{bai2023qwen}, and GPT-4V.
The growing interests in applying LLMs to specialized domains have also led to various innovations.
For instance, RadFM \cite{wu2023generalist} employs a LMM to the radiology and supports 3D CT and MRI medical image inputs for the first time. RS-ChatGPT \cite{osco2023potential} utilizes a LMM with CoT \cite{DBLP:journals/corr/abs-2201-11903} method to handle various complicated interpretation tasks in the filed of remote sensing and achieves promising performance.

\label{sec:intro}
\begin{figure}[t]
  \centering
  \includegraphics[width=8.5cm]{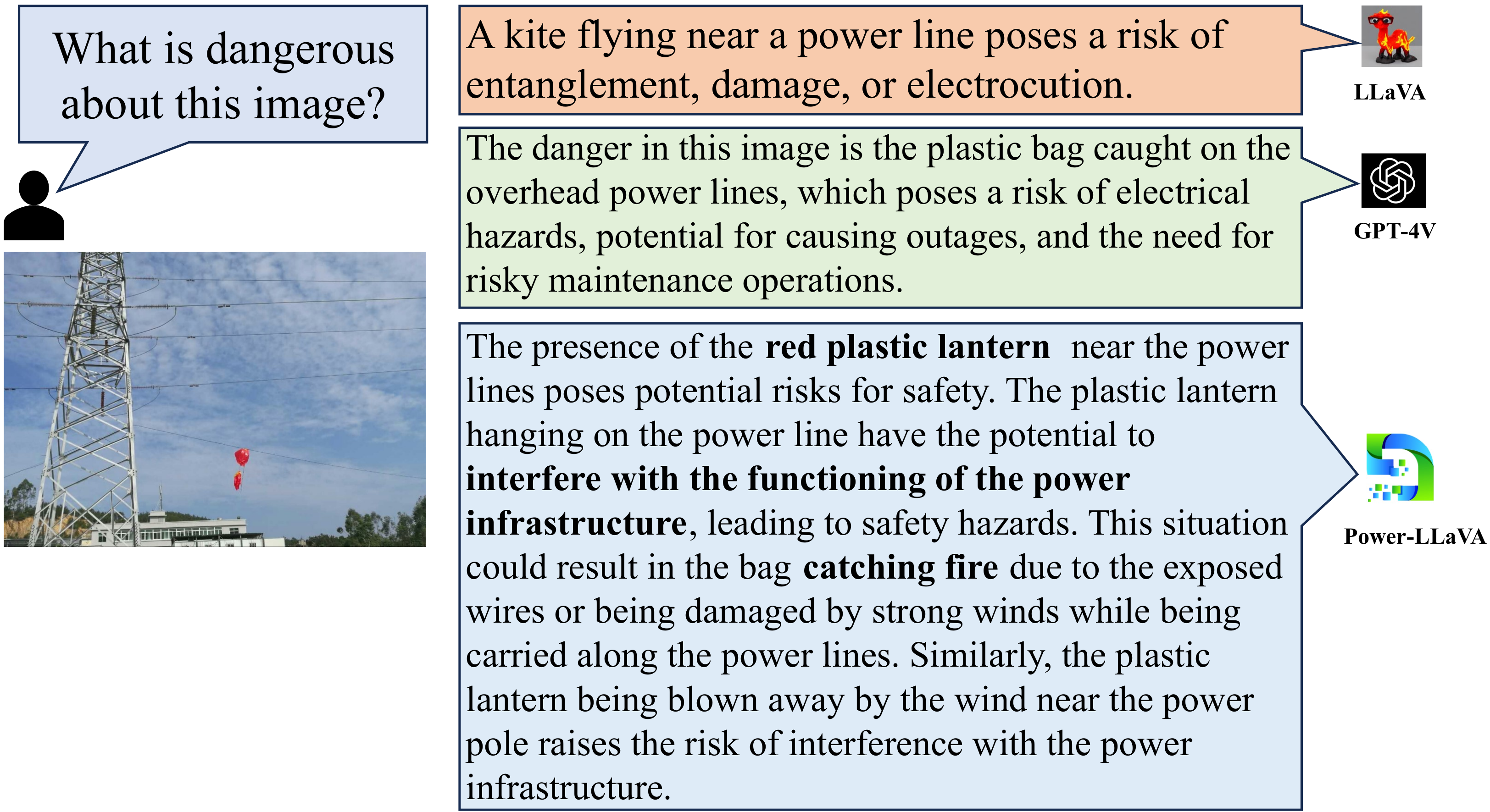}
  \caption{Comparison of  LLaVA, GPT-4V and Power-LLaVA. Power-LLaVA demonstrates the most comprehensive and specialized response towards power transmission line inspection.}
  \label{teser}
\end{figure}

 In this paper, we propose Power-LLaVA, the first intelligent vision-language  assistant based on the Large Multimodal Model (LMM) for power transmission line inspection through  interacting with humans. 
To accomplish this, we develop  a multi-modal model capable of accepting  image and text as inputs  and generating text response, as shown in Fig.~\ref{teser}.
The proposed model comprises of three  modules:  a pre-trained vision encoder (\emph{i.e.,} ViT \cite{dosovitskiy2020image}), a lightweight projection module, and a pre-trained LLM.
Specifically,  image patch embeddings are initially produced by the vision encoder and subsequently mapped into the  LLM's embedding space via the projection module.
The LLM module  processes both visual embeddings from the image and word embeddings from the text instruction in an unified manner, ultimately generating  text outputs as the response.

 To ensure that Power-LLaVA's response are well-aligned to the requirements of power transmission line inspection, we construct the large-scale, high-quality specialized dataset for two training stages.
 Initially, we collect  20K images from real-world power transmission  line  scenarios.
 Targeting at feature alignment in the first stage, we collect 608K brief image-text pairs, consisting of 558K from the public dataset released by LLaVA \cite{liu2023visual} and 50K generated by VL-models  \cite{bai2023qwen,huang2023language,liu2023visual}functioning as captioners.
The second stage is focused on instruction finetuning. Inspired by LLaVA \cite{liu2023visual}, we annotate each image with descriptive text and object detection labels by utilizing  existing advanced VL-models \cite{bai2023qwen,huang2023language,liu2023visual, chen2023sharegpt4v} and a detection model \cite{codetr2022}, respectively.
 To obtain a fine-grained instruction-following dataset, we employ ChatGPT  to generate  conversations that might occur during  transmission line inspections,  utilizing the captions and detection information derived from those inspection images.  
 In the end, our dataset in this stage comprises a total of 100K samples, providing a robust foundation for training and evaluating the Power-LLaVA model.

 Subsequently, We train our Power-LLaVA  applying a two-stage training strategy at a low computation cost. In the first stage, only the projection module is optimized to align visual embeddings with word emebddings of the LLM.
 During the second stage, both the projection module and the LLM are updated with the instruction-following dataset.
 Upon completion of  the two-stage training, our Power-LLaVA  achieves an accuracy of 86.79\%   on the  PowerQA benchmark, which consists of 1K samples generated by GPT-4V with manual checking and serves as a comprehensive measure of LMMs' inspection capabilities.
The performance of Power-LLaVA is highly competitive, even when compared to other state-of-the-art LMMs such as GPT-4V and Qwen.
\begin{figure*}[t!]
  \centering
  \includegraphics[width=15.1cm]{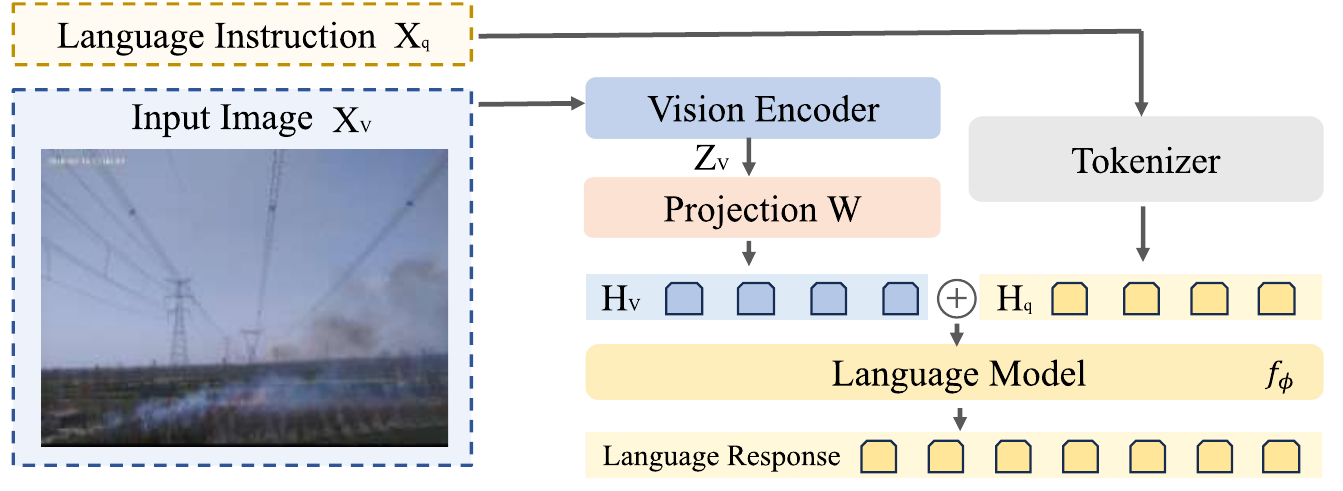} 
  \caption{Overview of our model. Initially, the vision encoder processes the input  image and extracts its feature as visual embeddings. These embeddings are then aligned with the word embeddings of the LLM via the projection module. Subsequently, the LLM module processes both the visual embeddings derived from the image and the word embeddings from the text in a unified manner, ultimately generating the text response.} 
  \label{fig:model_structure} 
\end{figure*}

In summary, our contribution has three folds:
\vspace{-0.6em}
\begin{itemize}

	\item We propose Power-LLaVA, the first intelligent vision-language  assistant based on the LMM for power transmission line.
 \vspace{-0.7em}
	\item We construct a large-scale and high-quality dataset specialized for the powerline inspection task.
 \vspace{-0.7em}
	\item We develop the PowerQA benchmark, which can comprehensively evaluate the model's ability to understand and perceive in power scenarios.
\end{itemize}

\section{Related Works}
\noindent \textbf{Large Multimodal Model}
The success of Large Language Models (LLMs) has attracted much research interests, engendering another research line focused on  extending the perceptual capacities of LLMs through the integration of other modalities.  
This is exemplifies by Flamingo \cite{alayrac2022flamingo}, which has demonstrated the notable ability of multimodal in-context learning, achieved  by harnessing a pre-trained visual encoder and a pre-trained large language model.
Similarly,  MiniGPT-4 \cite{zhu2023minigpt4} has exhibited  impressive conversational capabilities through pre-training for feature alignment and instruction tuning. 
LLaVA \cite{liu2023visual}, on the other hand, utilized a lightweight linear layer for feature alignment across modalities,
 contributing to superior  performance with an affordable training cost. Further research, such as ShareGPT4V \cite{chen2023sharegpt4v} and SVIT \cite{zhao2023svit}, has demonstrated the importance of training data for large multimodal models.

Simultaneously, the application of  LLMs to specialized domains has drawn a substantial amount of academic interests.
 BioMedGPT-LM \cite{luo2023biomedgpt} unified the feature spaces of molecules, proteins, and natural language through encoding and alignment, thereby enabling a performance  comparable to humans in biomedical question-answering tasks.
 Additionally, DocPedia \cite{feng2023docpedia} has achieved robust document understanding capabilities by extracting  visual features directly from the frequency domain as opposed to the pixel space.  
 FengWu \cite{chen2023fengwu} excelled in global medium-range weather forecasting, supported by 39-year-of training data based on the ERA5 reanalysis and a cross-modal fusion transformer.

\vspace{0.5em}
\noindent \textbf{Power Transmission Line Inspection}
Utilizing advanced deep learning techniques, substantial progress has been achieved in the domain of power transmission line inspection.
Zhang et al. \cite{zhang2020multi} introduced  a Multi-Scale Fusion Feature Alignment (MSFA) module and a Multi-Scale Consistency Regularization (MSCR) module to enhance the ability of Faster R-CNN \cite{ren2016faster} within this domain. 
The CSSAdet \cite{10251607}, which incorporated spatial and cross-scale attention mechanisms, has exhibited exceptional 
 performance in identifying  foreign objects around power lines. 
 Wan et al. \cite{9084729} employed Deformable Convolution and Squeeze-and-Excitation (SE) blocks, thereby enhancing  the R-FCN's \cite{dai2023rfcn} ability to exploit  fine-grained contexts within images.
 Liu et al. \cite{souza2023hybrid} proposed an algorithm that improved the overall learning capacity of YOLOX-S \cite{ge2021yolox} and increased detection accuracy by aggregating spatial and channel information in the feature map.
Nonetheless, current methodologies encounter difficulties in  generalization and intelligence, leading to suboptimal performance in complex application scenarios. 
\begin{figure*}[t!]
  \centering
  \includegraphics[width=17cm]{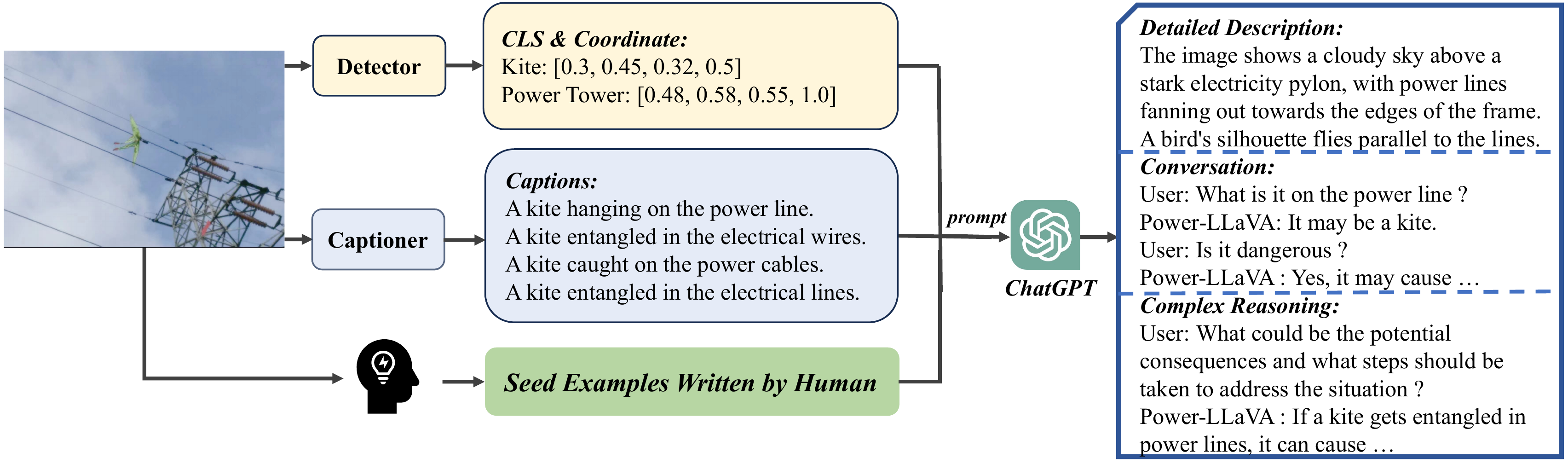} 
  \caption{Construction pipeline of our proposed dataset. For each image obtained from real-world power transmission line scenarios, we annotate four captions and object detection labels by utilizing state-of-the-art Vision-Language (VL) models and detection models for each image, respectively. Building upon the captions, object detection labels, and  templates provided by human annotators, ChatGPT is employed to generate a specialized high-quality dataset for instruction tuning.} 
  \label{fig:datagenerate} 
\end{figure*}

\section{Method}
\label{sec:method}

In this section, we will systematically illustrate the implementation  of our power transmission line inspection framework, denoted as  Power-LLaVA.
We initially present a comprehensive description of the architecture  of Power-LLaVA model.
Following this, we propose an effective pipeline capable of semi-automatically generating high-quality dataset. 
Finally, we provide our training strategy for Power-LLaVA, which comprises a two-stage training process.


\subsection{Model Architecture}

As illustrated in Fig.~\ref{fig:model_structure}, our Power-LLaVA model comprises three modules: a pre-trained vision encoder $ViT$, a lightweight projection module $Proj$, and a pre-trained LLM $f_{\mathrm{\phi}}$. Given an image $X_v$, the vision encoder first produces visual embeddings $Z_v$. The embeddings are then aligned with the word embeddings of the LLM via the projection module, forming the image tokens $H_v$ as:

\begin{equation}
{H_v} = Proj\left( {ViT}({X_v}) \right)
\label{for:vit}
\end{equation}
Meanwhile, the text instruction provided by humans is processed by a tokenizer to establish text tokens $H_q$. The image tokens and text tokens are then concatenated to form the input to the LLM which generates the text response:

\begin{equation}
{X_{response}}=f_{\mathrm{\phi }}(H_v,H_q)
\label{for:generate}
\end{equation}

\subsection{Dataset Construction}
\label{sec:data_gen}
\noindent \textbf{Dataset for Feature Alignment }
Initially, we construct a dataset comprising image-text pairs for the purpose of model pre-training.
This is achieved by amassing 20,000 images that depict real-world power transmission line scenarios, and subsequently annotating these raw images with brief descriptions and object detection labels using  advanced Vision-Language (VL) models and a detection model.
Specifically, we utilize VisualGLM \cite{liu2023visual}, InternLM-XComposer-VL \cite{zhang2023internlmxcomposer},  Qwen-VL-Chat \cite{bai2023qwen} and ShareGPT4V \cite{chen2023sharegpt4v} to generate captions, while Co-DETR \cite{codetr2022} serves as the detector.
Consequently, we generate 50K image-text pairs of power transmission line scenarios. 
\vspace{0.5em}

\noindent \textbf{Dataset for Instruction Tuning }
Drawing inspiration from LLaVA \cite{liu2023visual}, we develop an efficient data generation methodology to create a high-quality, specialized dataset, as depicted in Fig.~\ref{fig:datagenerate}.
Given that the data required for instruction tuning necessitates more fine-grained information and  diversity,
 we incorporate  ChatGPT to augment the data produced through the aforementioned process. 
 In order to increase diversity, we generate three types of conversation data: \emph{detailed description}, \emph{long conversation}, and \emph{complex reasoning}. Data samples belonging to the detailed description type consist of a single round of dialogue, primarily involving a question and the corresponding detailed description of the image. Long conversation samples encompass multi-round interactions between humans asking questions about the image and the assistant providing answers as if observing the image. Complex reasoning data focuses on samples that require reasoning about the visual content.

Furthermore,  we construct 100 conversation templates as seed samples for each data type.
Leveraging the powerful in-context learning capabilities of ChatGPT, a large-scale dataset for these three data types  could be generated by guiding ChatGPT with  randomly selected seed samples and appropriate prompts.
As a result, we obtain a specialized  dataset comprises of a total of 100K samples, which includes  22K in detailed descriptions, 45K in conversations and 33K in complex reasoning, respectively.


\subsection{Training Objective}
Power-LLaVA is optimized by  performing standard GPT-like training on the conversation dataset generated in Sec.~\ref{sec:data_gen}.
To be specific, all data is reformulated into the instruction-following manner by treating all questions and answers as instruction from the human and response from the assistant, respectively.
The $t$-th \textbf{$Instruction$} in the conversations with $T$ rounds is defined as 
\begin{equation}
Instruction^t = 
\begin{cases} 
([X,\,Q^1] \,\ \text{or} \ \, [Q^1, \, X]), & \text{if } t = 1 \\
Q^t, & \text{otherwise}
\end{cases}
\label{equ:instruction}
\end{equation}
where $X$ is the given image in current conversation and $Q^t$ is the question in the $t$-th round.

With the special system prompt designed to illustrate the assistant's functionality and responsibility, we could transform all training samples into an unified form of multimodal sequence as follows:

  \vspace{0.2em} 
\begin{center}
     \begin{tabular}{cc}
        \hline
        \textbf{Role} & \textbf{Content} \\
        \hline
        System & System prompt \textless \textbf{STOP}\textgreater \\
        Human & Instruction$^1$ \textless \textbf{STOP}\textgreater \\
        Assistant & \textbf{\textit{Response$^1$}} \textbf{\textless STOP\textgreater} \\
        Human & Instruction$^2$ \textless \textbf{STOP}\textgreater \\
        \multicolumn{2}{c}{...} \\
         Assistant & \textbf{\textit{Response$^T$}} \textbf{\textless STOP\textgreater} \\
        \hline
    \end{tabular}
    \vspace{0.2em} 
\end{center}  

\noindent Auto-regressive training objective thus can be employed on the unified sequence with a length of $L$:
\begin{equation}
    p(R|Instruct, X) = \prod_{i = 1}^{L}p_\theta(r_i|X, Instruct_{<i}, R_{<i})
    \label{equ:train}
\end{equation}
where $\theta$ represents the parameters to be optimized, and $Instruct_{<i}$, $R_{<i}$ denote the instruction and response tokens before current index $i$, respectively.

\subsection{Training Strategy}

We optimize our Power-LLaVA using a two-stage training strategy, encompassing an initial pre-training stage and a subsequent instruction tuning stage.

\vspace{0.5em}
\noindent \textbf{Stage 1: Pre-training for Feature Alignment}
The first stage aims to  align visual embeddings with word embeddings of the LLM.
In this stage, only the projection module is trainable, while both the vision encoder and the LLM module remain frozen.
This configuration enables the projection module to achieve optimal alignment performance while significantly reducing training  costs. 
During this stage, only image-text paired data is used for training, which is reformulated as single-turn dialogues. 
To maintain the general ability of our assistant, we also mix our generated data with   558K samples released by LLaVA.
\vspace{0.5em}

\noindent \textbf{Stage 2: Instruction Tuning}
In the second stage, we maintain the parameters of the  vision encoder frozen and continue to optimize the pre-trained weights of both the projection module and the LLM module in the end-to-end method.
During the second stage, we utilize the the instruction-following dataset constructed above.


\begin{table}[t]
\centering
\small
\caption{ Accuracy  on PowerQA dataset for different models. Power-LLaVA exhibits outstanding performance at a relatively low training cost.}

\begin{tabular}{@{}lccc@{}}

\toprule
Model           &Data scale           & Params         & Accuracy (\%)   \\ \midrule
Power-LLaVA     &  708K           &7B          & \textbf{86.79}     \\
VisualGLM              &  330M             &6B          & 62.27       \\
InstructBLIP           &  7.68M            &7B          & 32.54     \\
MiniGPT4               &   5M              &7B          & 63.43      \\
InternLM-XComposer-VL  &   $>$30M           &7B         & 81.08         \\
Qwen-VL-Chat        &  1.5B              &7B          & 84.45   \\ 
GPT-4V               &   -              & -       & 85.11  \\ \bottomrule
\end{tabular}
\label{label:label-1}
\end{table}

\section{EXPERIMENTS}
\label{sec:pagestyle}

\subsection{Evaluation Benchmark}
\label{sec:eval}

To comprehensively evaluate the capabilities of Power-LLaVA in the task of power transmission line inspection, we propose a GPT-4V-assisted PowerQA evaluation benchmark. This benchmark provides a metric for comparing our model against other advanced LMMs  by employing a zero-shot testing approach on the PowerQA dataset.

Initially, we instruct GPT-4V to generate relevant questions and answers from various perspectives based on real-world images of power transmission line inspection scenarios. These perspectives include the functionality of electrical equipment, potential hazards near cables, the surrounding environment of the equipment and so on. We structure these question-answer pairs as multiple-choice questions, offering two to four options for each question. Our PowerQA dataset consists of 1000 samples totally. Furthermore, to reduce the likelihood of GPT-4V generating samples with dirty issues, such as ambiguous options, misinterpretations of images, and repetitive questions, all samples are refined manually. This human intervention is crucial for the integrity and quality of the evaluation benchmark.

The evaluated models are instructed to directly select an answer for each question from the provided candidate options. Answers generated by the models that match the reference option are considered correct. We use accuracy as the criterion metric for evaluation.

\subsection{Setup}

To initialize the model,  the pre-trained ViT-L/14 \cite{radford2021learning} is employed as the vision encoder and vicuna-7B \cite{peng2023instruction} is utilized as the LLM module. Furthermore, we utilize a randomly initialized linear layer as the projection module for the purpose of striking an optimal balance between complexity and effectiveness.

In the pre-training procedure, we set the learning rate as 2e-4 and assign a batch-size of 4 for each GPU.
During the instruction fine-tuning stage, we adjust the learning rate and batch-size for each GPU as 2e-5 and 2, respectively. We employ an Adamw optimizer, a cosine learning rate scheduler, and  mixed precision strategy of float16 across both training stages. All experiments are conducted on 16 V100 GPUs.

\begin{figure*}[t]
  \centering
  \includegraphics[width=0.9\textwidth]{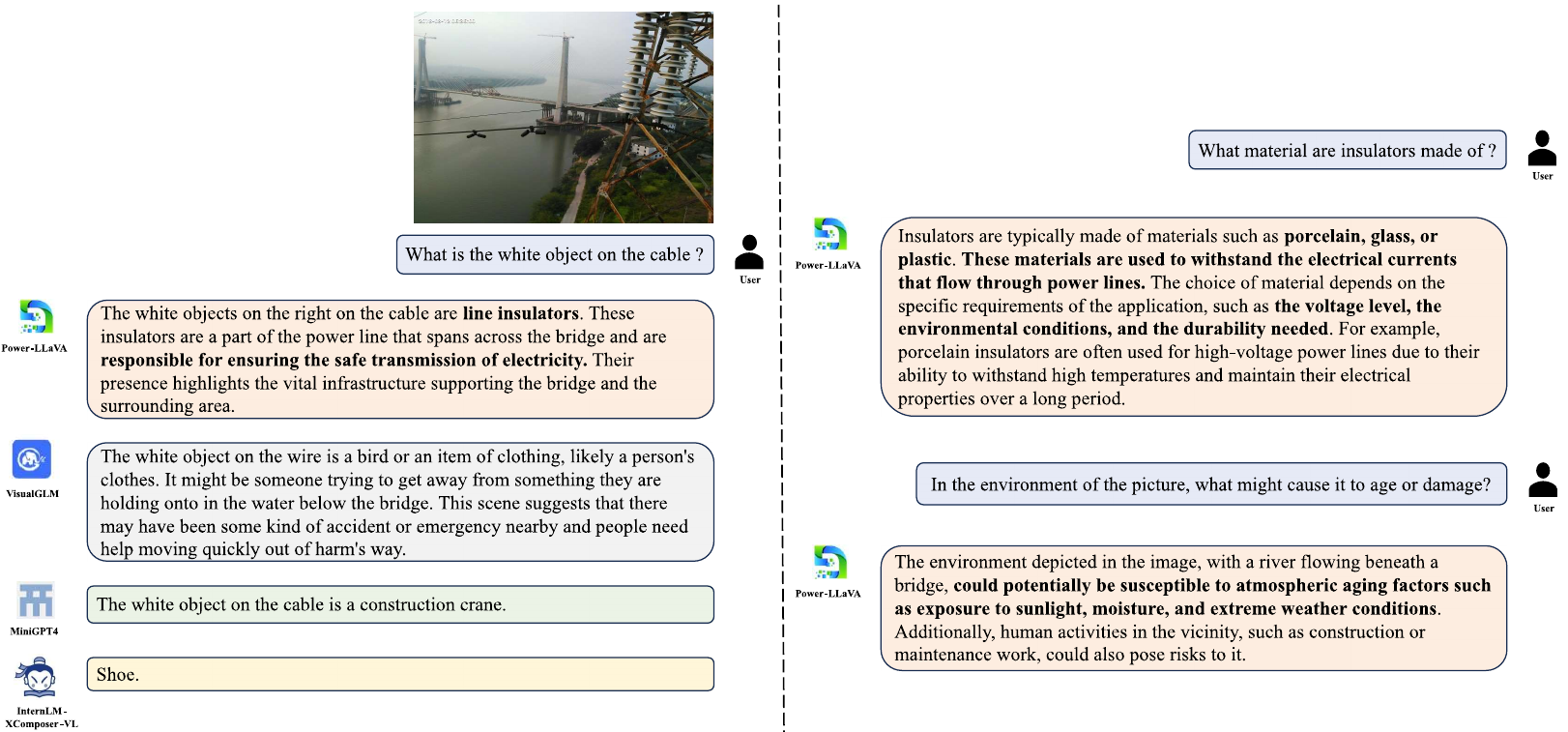} 
  \caption{Example of superior visual understanding and reasoning capability of Power-LLaVA in comparison to other models.  Power-LLaVA exhibits the ability to interpret images and instructions with a high level of professionalism. Additionally, it is capable of executing multi-round dialogues and complex reasoning tasks.} 
  \label{example} 
\end{figure*}

\begin{figure}[t]
  \centering
  \small
  \includegraphics[width=0.43\textwidth]{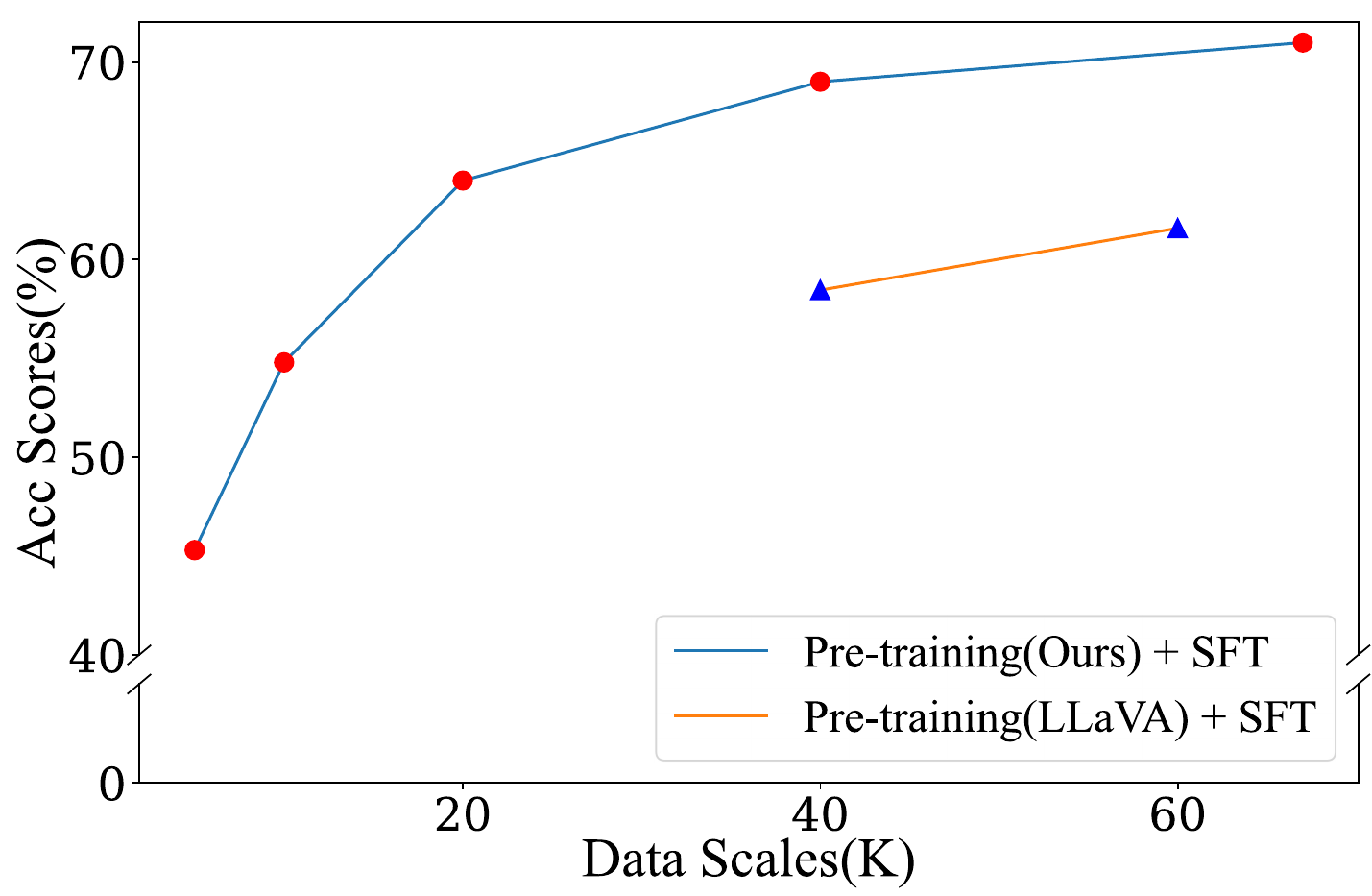} 
  \caption{The performance of our Power-LLaVA and LLaVA  when varying the scale of instruction finetuning dataset.} 
  \label{fig:res} 
\end{figure}

\begin{table}[t]
\centering
\caption{Accuracy on PowerQA with different training data.}

\begin{tabular}{@{}c  c  c | l@{}}
\toprule
Conv      &Detail     &Complex      & Accuracy (\%)     \\ \midrule
\Checkmark  &\Checkmark &\Checkmark  & 86.79       \\ 
\Checkmark  &\Checkmark &    &67.42 \textcolor{red}{$_{(-19.37)}$} \\  
\Checkmark  &  &\Checkmark  &49.83 \textcolor{red}{$_{(-36.96)}$} \\ 
   &\Checkmark &\Checkmark  &35.49\textcolor{red}{$_{(-51.30)}$} \\ \bottomrule

\end{tabular}
\label{datatype}
\end{table}

\subsection{Main Results}

Power-LLaVA and other competitive models are assessed on the PowerQA benchmark,  with a comparative analysis of these models presented in  Tab.~\ref{label:label-1}. 
The results indicate  that Power-LLaVA achieves the best performance on the PowerQA benchmark, surpassing other advanced LMMs such as GPT-4V, InternLM-XComposer-VL and Qwen-VL-Chat in response accuracy.
This supports the exceptional performance of our Power-LLaVA in the field of power transmission line inspection.

To  provide a more comprehensive evaluation, we include the  challenging examples in Fig.~\ref{example} that necessitate profound  understanding and reasoning capabilities. As Fig.~\ref{example} shows, Power-LLaVA can not only interpret images and instructions professionally but also exhibit in-depth reasoning,  which is critical for practical applications. 

Remarkably, in contrast to other LMMs, Power-LLaVA demands significantly fewer training data and computational resources.
While the data scale of LLM candidates reaches a minimum of 5 million, Power-LLaVA merely requires  708K.
Although Qwen-VL-Chat's performance is only marginally inferior to our  Power-LLaVA, its data size is five orders of magnitude lager than that of  Power-LLaVA.
In summary, Power-LLaVA  exhibits   exceptional  performance at a relatively low training expense. 

\subsection{Ablation Studies}
In this part, we conduct a series of ablation studies to dissect the contributions of key components in our method, including the scale of dataset used for training, the composition of the dataset, and the two-stage training strategy.

\vspace{0.5em}
\noindent \textbf{Impact of Data Scale on Model Performance}  As shown in Fig.~\ref{fig:res}, the accuracy score rises up rapidly when the size of dataset is smaller than 40K and then gradually plateaus at 60K. Therefore, 100K is set as the final size of our dataset for the purpose of balancing the performance and training cost.

\vspace{0.5em}
\noindent \textbf{Importance of Pre-training with Specialized Data}
We conduct pre-training using our specialized dataset as part of the two-stage training process. To assess its importance, we compare this approach with LLaVA's pre-training, which utilized open-source dataset. 
As shown in Fig.~\ref{fig:res}, LLaVA's pre-training leads to inferior performance in the power domain compared to our specialized pre-training approach, which increases the average accuracy by approximately 10\%. We hypothesise that the performance gap may be due to the non-negligible domain gap between the dataset used by LLaVA and our target scenarios.

\vspace{0.5em}
\noindent \textbf{Different Types of Data}
We examine the impact of various data types in our tuning dataset by excluding one type at a time during training.
As shown in Tab.~\ref{datatype}, removing any of the three types of data results in  obvious degradation in model performance, suggesting that all three types of data are of great significance. 
Additionally, the data belonging to complex reasoning seems to be the least important, while the conversation data is the most critical. 

\section{Conclusion}
\vspace{-0.2em}
We propose Power-LLaVA, the first large language and vision assistant specially designed for power transmission line inspection. 
This framework integrates a pre-trained vision encoder with a large language model, thereby promoting a promising inspection capability through an efficient  two-stage training strategy.
Furthermore, we construct the first specialized dataset and comprehensive benchmark within this field.
As for limitations, the effectiveness of our method has not been substantiated via scaling up either model capacity or data scale.
It is our aspiration that  our work will stimulate further research in application  of electrical engineering.
\vspace{0.2em}

\noindent\textbf{Acknowledgement} This work was supported by National Natural Science Foundation of China (No. U2166214). 


\label{sec:refs}

\bibliographystyle{IEEEbib}

\end{document}